# Decentralized, Self-organizing, Potential field-based Control for Individually-motivated, Mobile Agents in a Cluttered Environment: A Vector-Harmonic Potential Field Approach.

By
Ahmad A. Masoud
P.O. Box 287, Electrical Engineering Department, King Fahd University of Petroleum and Minerals, Dhahran 31261, Saudia Arabia,
E-mail: masoud@kfupm.edu.sa, Web: http://faculty.kfupm.edu.sa/EE/masoud/, Tel: (03) 860-3740.

## Abstract

Spatial multi-agency has been receiving growing attention from researchers exploring many of the aspects and modalities of this phenomenon. The aim is to develop the theoretical background needed for a multitude of applications involving the sharing of resources by more than one agent. A traffic management system is one of these applications. Here, a large group of mobile robots that are operating in communication-limited, and sensory-limited modes are required to cope with each others presence as well as the contents of their environment while preserving their ability to reach their preset, independent goals. This work explores the construction of a decentralized traffic controller for a large group of agents sharing a workspace with stationary forbidden regions. The suggested multi-agent motion controller is complete provided that a lenient condition on the geometry of the workspace is upheld. It has a low computational effort that linearly increases with the number of agents. The controller is also self-organizing; therefore, it is able to deal, on its own, with incomplete information and unexpected situations. In addition to the above, the controller has an open structure to enable any agent to join or leave the group without the remaining agents having to adjust the manner in which they function. To meet these requirements, a definition of decentralization is suggested. This definition equates decentralization to self-organization in a group of agents operating in an artificial life mode. The definition is used to provide guidelines for the construction of the multi-agent controller. The controller is realized using the potential field approach.  Theoretical developments, as well as simulation results, are provided.

# I. Introduction

The scarcity of resources in modern environments makes it necessary for the agents occupying that environment to share available resources. Whether it is the congested airspace of international airports, roads and highways at rush hours, or the busy downtown sidewalks of a metropolitan city, the agents must make intelligent use of the resource of space for each to safely reach its target, hopefully, along the shortest path that the situation permits. Multi-agent systems are the focus of intensive investigation by researchers and engineers [52,53]. The main goal of research in this area is to conserve resources via efficient utilization, and/or to tackle large tasks cooperatively. This goal is seriously hindered by problems that can arise as a result of two or more agents attempting to utilize a resource in conflicting ways (e.g. trying to occupy the same space at the same time). Designing social agents is a difficult task. This is due to the simple fact that coexisting in an environment changes the nature of the agents from that of individuals into that of interconnected members of a group affected by each other's actions. An improper action on the part of an agent can directly (by harming other agents) or indirectly (by failing to fulfill a role that is vital to others) have an adverse effect on other agents. The affected agents may not necessarily be in the immediate physical proximity of the offending agent (a chain effect). To shed some light on the difficulties encountered by a multi-agent system, consider a daily act of planning which people engage in with little, if any, attention to its complexity. The act is the simple trip from home to work and back. In a metropolitan city such a process involves thousands if not millions of participants, each of whom is only aware of his/her destination. The hard-to-acquire information about the constituents of the environment and the intentions of the other agents is not expected to be of much help. Any attempt to use this information to derive *a priori* known, conflict-free, goal-oriented trajectories will face serious difficulties. In societies of individually motivated agents, communication costs are prohibitive [56]. As for the intellectual labor needed to manage such a process in order to avoid conflict and guarantee that each agent will safely reach its target, each path has to be checked, along with goal satisfaction, for possible conflict with the remaining N-1 paths of the other agents (N agents are assumed to be participating in the above process). This is highly likely to translate into an exponential complexity ($N^N$) that is more than enough, on its own, to cripple any attempt of a central controller to coordinate the behavior of such a large group. It is not difficult to extrapolate the actual level of difficulty a realistic, large scale, multi-agent system faces. In such a situation no *a priori* considerations are given, or, in the opinion of this author, can be given to whether a path selected by an agent conflicts with the ones selected by others. The agents are highly unlikely to have *a priori* knowledge of all or any of the agents sharing their environment, let alone knowing their intentions. Amazingly, such a massive, purposive, organizational system seems to almost always operate well in the face of incomplete information and the perceived need for highly intensive computational requirements.

The above example is just one facet of multi-agent systems. Spatial multi-agency has been applied in air traffic [1-3], and vehicular traffic [4] management systems, industrial assembly [5], computer game design [6], mapping [7], and automated reconnaissance systems [8]. Understandably, the literature abounds with work on the theoretical foundation of individually motivated multi-agent systems, with coordination and conflict management [9-12] being the central topic of investigation. Although an intuitive assessment of the computational demands such systems may require has been supplied at the beginning of this section, it has been theoretically shown that the general multi-agent problem is PSPACE-complete [12,6]. This proves that the complexity of searching for a solution may have a lower bound that is exponential in the number of agents.

There are two main focal points in the study of mobile multi-agents: the agents are either viewed as a collective whose motion is motivated by a single group goal (group-motivated), or they are viewed as individuals each motivated by its own goal (individually-motivated). In the first type motion of agents as a team or a flock is studied with emphasis on deriving inter-agent coordination mechanisms for constructing adaptive spatial formations. Several approaches were considered for such a task. McInnes [13] and Schnider et al. [14] used the potential field approach for constructing a group navigator. Graph-theoretic techniques treating a flock as a spatially-induced graph were examined in [15-17]. Distributed nonlinear control schemes that are able to coordinate the behavior of the agents in a manner that would give rise to complex formation maneuvers by the group were suggested in [18-20]. Other approaches to constructing formations using self-organization, heuristic-reactive, and qualitative techniques may be found in [21-22] respectively. A method utilizing the motor schema approach for such a purpose [65] may also be found in [23,66].

Purposive agents having independent goals may be divided into two main classes. The first class is that concerned with planning motion for one agent only that is sharing its workspace with a non-cooperative group of agents. In this scenario the rest of the agents do not reactively adjust their paths to accommodate the presence of the agent concerned. Techniques dealing with such a situation may be found in [24- 28]. The second class presents a cooperative scenario where all agents simultaneously participate in reaching an accommodating arrangement that enables all of them to reach their respective destination. Cooperative planning techniques are, in general, more difficult to design than non-cooperative ones. Examples of cooperative methods may be found in [29-43]. Two main approaches for the construction of such planners are based on geometry or potential fields. While provably-correct, geometry-based methods that can effectively handle up to a medium size group do exist [29-34], in general, the prevalently centralized nature of such techniques results in an exponential complexity that makes their use with a large number of agents undesirable. Potential field methods [35-39], on the other hand, may be easily configured in a decentralized mode. The advantages of decentralization are numerous, some of which are: low complexity, high reliability and adaptability. While many approaches were explored for building decentralized planners [39-43], multi-agent, decentralized, planning is still a challenge. The main two issues in these approaches seems to be: proving the correctness of the planning procedure and better control over complexity growth and the process of factoring the influence of context and constraints in the behavior generation process. In the few cases where a provably-correct decentralized planner was suggested a restrictive view of decentralization was adopted. For example, in [39] a provably-correct, potential field-based, multi-agent planner was proposed. However, decentralization was considered only in the sense of each system having no knowledge of the targets of the other systems.

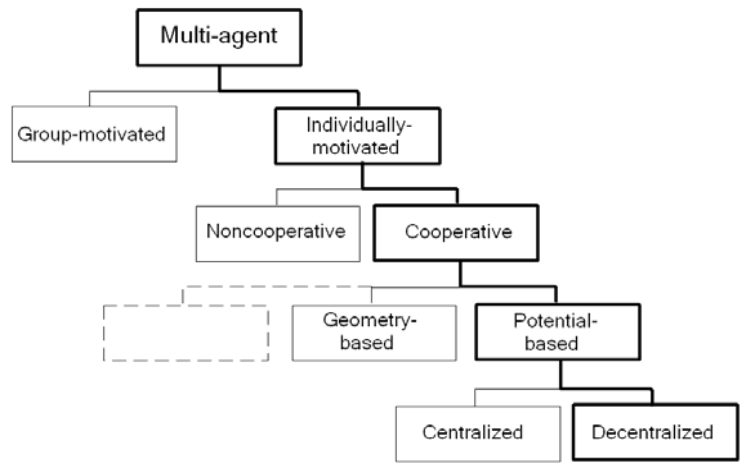

Figuer-1: A partial taxonomy of multi-agent systems.

The focus in this work is on constructing a decentralized, potential field-based, cooperative, individually-motivated, multi-agent motion planner (a partial taxonomy of multi-agent systems is shown in figure-1. Work attempting to classify multi-agent systems may be found in [44,45]). The planning problem tackled here is a practical, special case of the general spatial, multi-agent planning problem that imposes no constraints on the environment. Special attention is paid to developing a definition for decentralization capable of supporting the construction of a planner with the following properties:

1-provably-correct and complete, provided that a lenient condition on the geometry of the workspace is upheld (i.e. if a solution exists provided that the condition is upheld, the planner will find it; otherwise, it indicates that the problem is insolvable);

2-flexible (i.e. the event of agents joining or leaving the group will not necessitate that each member of the collective accommodate this change in the method it uses to generate actions. Only the agents physically proximate to where the change occurred have to carry out such an adjustment. To the rest of the collective, the agents newly arriving or departing remain transparent) ;

3-fault tolerant (i.e. if during operation one or more agents unexpectedly fail, the remaining agents will still be able, with a high probability, to continue unaffected to their targets);

4-computationally feasible for a large group-the planner suggested here has linear complexity in the number of agents;

5-functional in informationaly-deprived situations where the static environment need not be *a priori* known; also the agents need not *a priori* know each other.

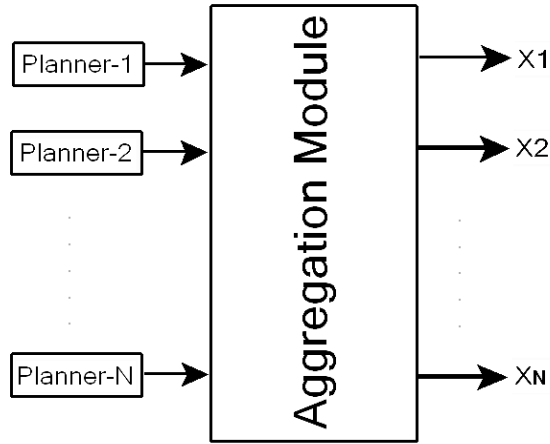

Figure-2: Layout of the suggested, multi-agent planner

The planner is divided into two stages (figure-2):

1- a stage that consists of N single-agent planners each of which is acting independently of the others as if it were the only active entity utilizing the workspace. Each one of these planners is referred to as the purpose field controller (PRF). Their task is to steer the corresponding agent, in a constrained manner, to *it's a priori* specified target.

2- an aggregation module whose primary function is decentralized conflict management. This module intervenes only when a conflict situation is in close proximity to an agent. It temporarily modifies the guidance actions from the PRF so that the agent is steered along a conflict-free path during that period. The conditioning action from this module quickly dissipates after the conflict is resolved, giving back full control over motion to the PRF component. This module is called the conflict resolving field (CRF) control.

The N single-agent planners (PRF controllers) used for building the multi-agent planner are constructed using an existing approach which the author participated in developing. The approach employs evolutionary, hybrid, pde-ode controllers (EHPCs) which are constructed using potential fields that are set in an artificial life (AL) mode. The general framework for such a type of planners was presented in [46]. Different realizations of this framework may be found in [47-50].

The main contribution of the paper lies in the construction of an evolutionary aggregation module that conforms to the guidelines of AL [51]. The module operates in the manner described above, is provably-correct, and, most importantly, the effort it exerts for guidance and conflict mediation is linear in the number of agents. The module is designed for working with EHPCs. The conflict resolving action modifier it employs is constructed from an underlying vector potential field. As was demonstrated in [50], an action generated from an underlying vector potential has superior motion steering capabilities compared to one generated from an underlying scalar potential.

This paper is organized as follows: section II of the paper discusses centralized and decentralized control. It also provides an interpretation of decentralization using the artificial life approach to behavior synthesis. The multi-agent motion control problem is formulated in section III. A realization of the controller is suggested in section IV. Section V provides an analysis of the controller's ability to safely drive each agent to its target. The behavior of the suggested controller is explored using simulation experiments in section VI, and conclusions are placed in section VII.

## II. Centralized Versus Decentralized Control

In this section the general properties of centralized control systems are briefly presented. A definition of decentralization that is derived from self-organization in a collective of agents set to operate in an AL mode is proposed for the multi-agent case. With the help of the potential field approach, the definition is used in section IV to realize the CRF and PRF control components used in constructing the multi-agent controller. A general discussion of multi-agent systems along with a comparison between centralized and decentralized control may be found in [52,53] and [54,55] respectively.

A: Centralized, multi-agent systems:

Whether it involves one or more agents, successful, context-sensitive, purposive behavior requires the presence of a process for generating a regulating control action . This process receives data from the environment, the agent(s), the target(s), and the constraints on behavior, and converts them into a control action that should successfully propel the agents, in a constrained manner, towards their goals. There are two ways for generating such a regulating action: a centralized approach, and a decentralized approach.

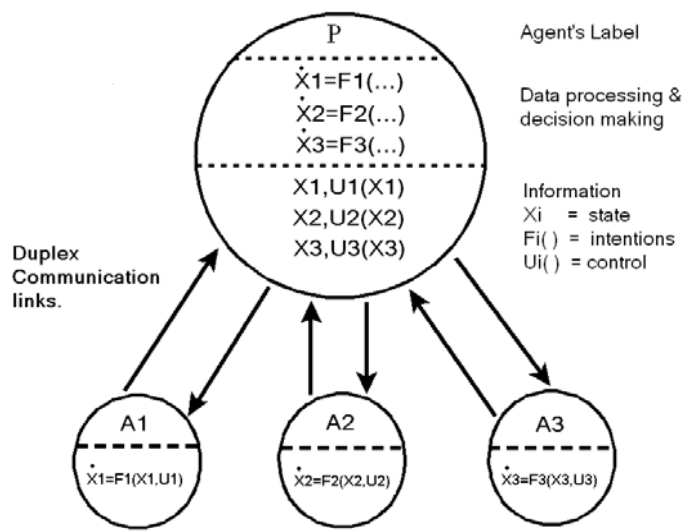

Figure-3: Centralized approach to control

The centralized approach has a holistic-in-nature, top-down view of the behavior synthesis process. Here, a central agent that has a duplex communication link to each member of the group simultaneously observes the states of the agents and the environment, and processes the database in a manner that is in accordance with the aim of the group and the constraints on behavior. It then generates synchronized sequences of action instructions for each member. The instructions are then communicated to the respective agents for them to progressively modify their trajectories and safely reach their destinations (figure-3). In this mode of behavior, the generation of the constraint-satisfying, goal-fulfilling, conflict-free solution (i.e. sequence of state-control pair) begins by constructing the hyper action space (HAS) of the group. HAS contains the space of all admissible point actions which the agents may attempt to project. The HAS is then searched for a solution that is in turn communicated to the agents. The agents reflexively execute the solution trusting that their actions will lead to the desired conclusion. It is a well-known fact that, in real life, any solution generated by a centralized mechanism is short lived. The dynamic nature of real environments will cause a mismatch between the conditions assumed at the time the controller begins generating the solution, and the actual conditions at the time the solution is handed to the agents for execution. Despite the attempt to alleviate this problem by equipping the agents with local sensory and decision making capabilities, large scale, centralized systems still suffer serious problems, some of which are stated below:

▸ Almost all centralized planning and control problems are known to be PSPACE-complete with a worst case complexity that grows exponentially with the number of agents. The large number of agents a traffic system contains will prevent a central controller from adapting to environmental changes in a timely manner, if not crippling the control process altogether.

▸ Centralized systems are inflexible in the sense that any changes to the characteristics of one or more agents may translate into a change in the whole HAS. This makes it necessary to repeat the expensive search for a solution. In turn, the desirable property that the size of the effort needed to adjust the control be commensurate with the size of changes in the setting is not satisfied.

▸ Centralized systems are prone to problems in communication and action synchronization. This makes it difficult to reliably operate a large scale system even if the central planner has the computational assets needed to meet the demands of a realistic environment.

▸ Centralized systems are not robust in the sense that the failure of one agent to fulfill its commitment towards the group could lead to the failure of the whole group.

B: Decentralized multi-agent systems:
In real-life, no agent, no matter how sophisticated it is, has omniscient awareness of its surroundings, let alone infinite resources to instantly store and process data. Sometimes, even reliable communication links between the central agent and the others are difficult to establish. Communication may even be impossible due to the lack of a universally accepted technical language, even vocabularies. The above are a few reasons why central planning strategies may not succeed with real-life, large scale systems. Ruling out the feasibility of a central planner leaves only the option of the regulating control action arising from the agents themselves. The fact that the agents possess only local sensing, reasoning, and action capabilities makes it impossible to capture a complete spatial and/or temporal representation of the process. This, in turn, makes it impossible to build an HAS.

Obviously, it is not feasible for agents in a large group with distinct goals to be *a priori* aware of each other's presence, to communicate with each other or with a central agent regarding advice about what action to take. The only remaining option is for each agent to make its own decision on how to act based on the sensory data which the agent dynamically extracts from its local surroundings (Figure-4). Knowing that there is more than one interpretation of decentralization, the author considers a multi-agent system decentralized if each agent in the group is independent from the others in sensory data acquisition, data processing, and action projection. In a decentralized system, these faculties are configured in a mode that would give rise to coordination in the group without a coordinator. In other words, the group is capable of self-organization. Unlike centralized, top-down approaches, self-organization is a bottom-up approach to behavior synthesis where the system designer is only required to supply the individual agents with basic, "self-control" capabilities. The overall control action that shapes the behavior of the agents evolves in space and time as a result of the interaction of the agents between themselves and with their environment. Some properties of decentralized systems that conform to the above definition are:

▸No need to search or, for that matter, construct the HAS of the group in order to generate a solution. For a decentralized system, the solution emerges as a result of the agents interacting among themselves and with their environment.

▸No inter-agent communication, or communication with a supervisory agent. All that an agent is required to do is to observe (not communicate with) other agents in its local neighborhood. No preexisting awareness of the whole group, or the whole environment is required.

▸Synchronous behavior being an emergent phenomenon (instead of an imposed one) that results from asynchronous interaction.

▸The cost of computing the control in the group grows linearly with the number of agents.

▸Decentralized agents form open systems that enable any agent to join or leave the group without the others having to adjust the manner in which they process information or project action. This is a consequence of each member of the group being able to independently sense its environment, process data, and actuate motion.

▸ Unlike centralized systems which are informationally-closed, and organizationally-open, decentralized systems are informationally-open and organizationally-closed.

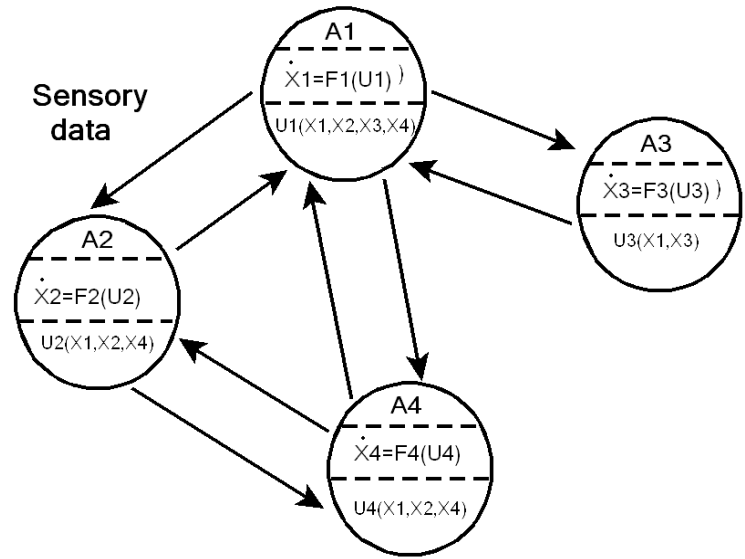

Figure-4: Decentralized approach to control

The difference between centralized and decentralized systems goes far beyond the manner in which the behavior generation faculties are related to the agents. It reaches as deep as the process enabling the system to generate the information needed for behavior synthesis. Centralized systems use reasoning coupled with search as the driver of the action selection process (it ought to be mentioned that function(al) minimization is a form of search). The search of the system's space of possible actions for a feasible solution may be carried out in a brute force manner, or in an intelligent manner that utilizes heuristics and side information for speed. No matter what form the search assumes or how it is applied, systems relying on search have problems (some of which are mentioned above) if they operate in a dynamic environment. On the other hand, the action selection driver in decentralized systems that satisfies the above requirements is a synergy-driven evolution. In this mode of behavior information synthesis is the result of the synergetic interaction of the agents among themselves under the influence of their environment. The information that is *a priori* encoded into each agent in the form of self-capabilities to project actions is usually simple and not adequate, on its own, to handle the usually complex planning task which faces the group. It is synergetic interaction within the context of the environment that augments the level of information of the group to a level that is sufficient for the members to carry out the task at hand (an act of knowledge amplification).

Artificial life (AL) [51] seems to provide a powerful paradigm for explaining the behavior of decentralized systems. It also provides constructive guidelines for their synthesis. In an AL system, the members of the group are equipped with the proper elementary, *a priori* known capabilities for self-control which are called the Geno-type of behavior (G-type). On the other hand, the overall control action that actually governs the behavior of the whole group evolves in space and time as a result of the interpretation of the G-type in the context of a particular environment (a process of morphogenesis [64]). The whole control action is called the Pheno-type (P-type) of behavior. This behavior cannot be exactly, *a priori*, predicted; only certain aspects of it can be *a priori* known. It is very flexible, highly adaptive, and far exceeds in complexity and informational content the G-type control. There are two requirements for constructing a proper G-type control action:

▸ Each agent must individually develop a control action to drive it toward its goal. Such a control need not take into consideration the control actions generated by the other agents of the group.

▸ Each agent must have the ability to generate a control that can resolve conflict with other agents through bilateral interaction.

## III. Formulation

In this section the problem of decentralized, multi-agent motion planning in the face of incomplete information is formulated. Here, an agent ($D_i$) is assumed to be massless, and occupy a set of points that forms a multi-dimensional, hyper sphere ($x \in R^M$) with a radius $\rho_i$ and a center $x_i$:

$$D_i(x_i, \rho_i) = \{x : |x - x_i| \leq \rho_i\} \qquad i=1,..,L, \qquad (1)$$

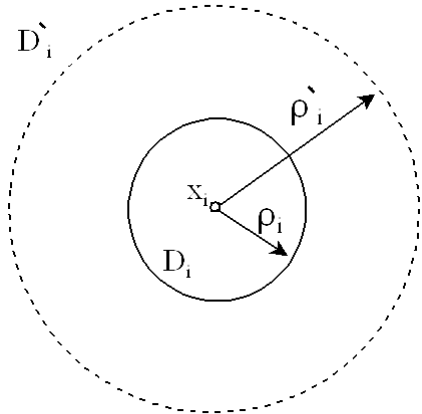

Figure-5: Zones related to the agent

where L is the number of agents occupying the workspace (figure-5). An enlarged circular region ($D\grave{}_i$) with radius $\rho\grave{}_i$ ($\rho\grave{}_i > \rho_i$) and center $x_i$ is assumed to be surrounding $D_i$:

$$D\grave{}_i(x_i, \rho\grave{}_i) = \{x : |x - x_i| \leq \rho\grave{}_i\} \qquad i=1,..,L, \qquad (2)$$

$$D_i \subset D\grave{}_i.$$

here and in the rest of the paper $D_i$ and $D\grave{}_i$ are used to refer to $D_i(x_i, \rho_i)$ and $D\grave{}_i(x_i, \rho\grave{}_i)$ respectively. The ring $S_i$ ($S_i = D\grave{}_i - D_i$) surrounding $D_i$ marks the region illuminated by the sensors of the i'th agent. The time between an agent sensing an event and releasing a control action (data processing and action release delay) is assumed small enough to be neglected. Therefore, this region is a dual sensory and action zone. Besides the agents, the environment is assumed to contain static, forbidden regions (O) which the agents must not occupy at any time ($O \cap D_i \equiv \phi, \forall t, i=1,..,L$). The agents are only allowed to exist in the workspace $\Omega$ ($\Omega = R^M - O$). The boundary of the forbidden regions is referred to as $\Gamma$ ($\Gamma = \partial O$). The destination of the i'th agent is surrounded by the spherical region $T_i$ with a center $C_i$ (figure-6). $T_i\grave{}$s are chosen so that:

$$\grave{D}_i \subset T_i \qquad x_i = C_i \qquad (3)$$
$$T_i \cap T_j \equiv \phi \qquad i \neq j$$
$$O \cap T_i \equiv \phi \qquad i=1,\ldots,L.$$

The last two conditions, respectively, mean that the goals of the different agents should not be conflicting, and should be attainable (i.e. lie inside $\Omega$). The partial knowledge the i'th agent has about its stationary environment is represented by $\grave{\Gamma}_i$ ($\Gamma \supseteq \grave{\Gamma}_i \supseteq \phi, i=1,\ldots,L$). The discrete-in-time, binary variable $Q_i$ ($Q_i \in \{0,1\}$) marks the event of a novel discovery of parts of a forbidden region, i.e.

$$S_i \cap \Gamma \neq \phi, \qquad (4)$$
and $\qquad ((S_i \cap \Gamma) - (S_i \cap \Gamma) \cap \grave{\Gamma}_i) \neq \phi \qquad i=1,\ldots,L.$

If at any instant in time ($t_n$), this condition becomes true, the content of $\grave{\Gamma}_i$ is adjusted so that:

$$\grave{\Gamma}_i(t_n) = \grave{\Gamma}_i(t_{n-1}) \cup (S_i \cap \Gamma). \qquad (5)$$

If such a situation transpires, $Q_i(t_n)$ is set to 1, otherwise, its value is set to zero. The i'th agent also actively monitors its immediate neighborhood for the presence of other agents. It forms the set:

$$\chi_i(t) = \{x = \bigcup_j D_j : S_i \cap D_j \neq \phi, j=1,\ldots, K_i(t), i \neq j\}, \qquad (6)$$

where $K_i(t)$ is the number of agents lying in the proximity of the i'th agent at time t. Designing the multi-agent controller requires the synthesis of the dynamical systems:

$$\dot{x}_i = h_i(x_i, C_i, Q_i, \chi_i, \grave{\Gamma}_i) \qquad i=1,\ldots,L, \qquad (7)$$

such that for the overall system: $\qquad \dot{X} = \mathbf{H}(X, C, Q, \Psi, \grave{\Gamma})$, $\qquad (8)$

$$\lim_{t \to \infty} x_i(t) \to C_i \qquad i=1,\ldots,L$$
$$D_i \cap D_j \equiv \phi \qquad \forall t, i \neq j$$
$$O \cap D_i \equiv \phi,$$

where $x_i \in R^M$, $X=[x_1 \ldots x_L]^T$, $C=[C_1 \ldots C_L]^T$, $Q=[Q_1 \ldots Q_L]^T$, $\grave{\Gamma}=[\grave{\Gamma}_1 \ldots \grave{\Gamma}_L]^T$, $\mathbf{H}=[h_1 \ldots h_L]^T$, $\Psi=[\chi_1 \ldots \chi_L]^T$.

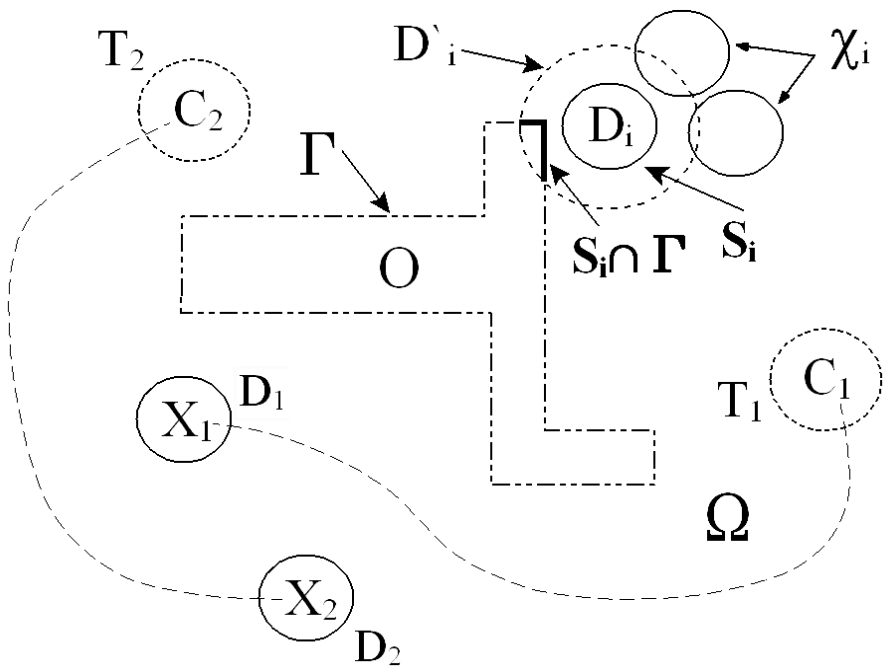

Figure-6: Goal oriented agents in a cluttered environment

## IV. Controller Design

As discussed earlier in the paper, adopting an AL approach to behavior synthesis reduces the job of the designer to the construction of only the self-controllers (G-type control) of the agents as individuals. The overall control action that regulates the behavior of the agents as a group operating in the context of some environment (P-type control) evolves as a result of the constrained synergetic interaction among the agents. The designer is required to synthesize controls for the systems:

$$\dot{x}_i = \mathbf{u}_i = h_i(x_i, C_i, Q_i, \chi_i, \Gamma_i^`) \qquad i=1,...,L. \tag{9}$$

The i'th self-control is divided into the following three components:

$$\mathbf{u}_i = \mathbf{ug}_i(x_i, C_i, Q_i, \Gamma_i^`) + \mathbf{uc}_i(x_i, \chi_i) + \mathbf{uo}_i(x_i, \Gamma_i^`), \tag{10}$$

where $\mathbf{ug}_i$ is the PRF component of the i'th self-control, $\mathbf{uc}_i$ is the CRF component, and $\mathbf{uo}_i$ is an optional control component that is included as an extra precaution against collision with stationary obstacles. $\mathbf{uo}_i$ is taken as the positive gradient of a potential field constructed as the inverse distance to the obstacle closest to a robot. In practice the potential is substituted for by a signal derived from a proximity range sensor. Details about how to construct $\mathbf{uo}_i$ may be found in [57]. It ought to be mentioned that $\mathbf{ug}_i$ includes, among other things, the ability to avoid collision.

### A. The PRF Control

The PRF controllers (self-controllers) are constructed using an evolutionary, hybrid, pde-ode control framework. This section provides only a brief overview of EHPCs. For a detailed discussion of EHPC, and a proof of correctness the reader is referred to [46], [47-50], and [49] respectively.

An EHPC (figure-7) consists of two parts:

1- a discrete time-continuous time system to couple the discrete-in-nature data acquisition process to the continuous-in-nature control action release process;

2- a hybrid, PDE-ODE controller to convert the acquired data into in-formation that is encoded in the structure of the micro-control action group.

The EHPC representing the i'th PRF control component is:

$$\mathbf{ug}_i = -\nabla V_i(x_i, C_i, Q_i(t_n), \Gamma_i^`(t_n)) \tag{11}$$

so that for the gradient dynamical system:

$$\dot{x}_i = -\nabla V_i(x_i, C_i, Q_i(t_n), \Gamma_i^`(t_n)), \tag{12}$$
$$\lim_{\substack{n \to Z \\ t \to \infty}} x_i(t) \to C_i \qquad i=1,...,L, \ n=1,...,Z$$

and: $\qquad D_i \cap O \equiv \phi \qquad \forall t,$

where n represents the n'th instant at which condition (4) becomes valid ($t_n$), Z is a finite, positive integer, and $\nabla$ is the gradient operator. At $t_n$, which marks the transition of $Q_i(t_n)$ from 0 to 1, first the contents of $\Gamma\grave{}_i$ are adjusted according to (5). The structure of the guidance field of the EHPC is then adjusted to incorporate the newly acquired data.

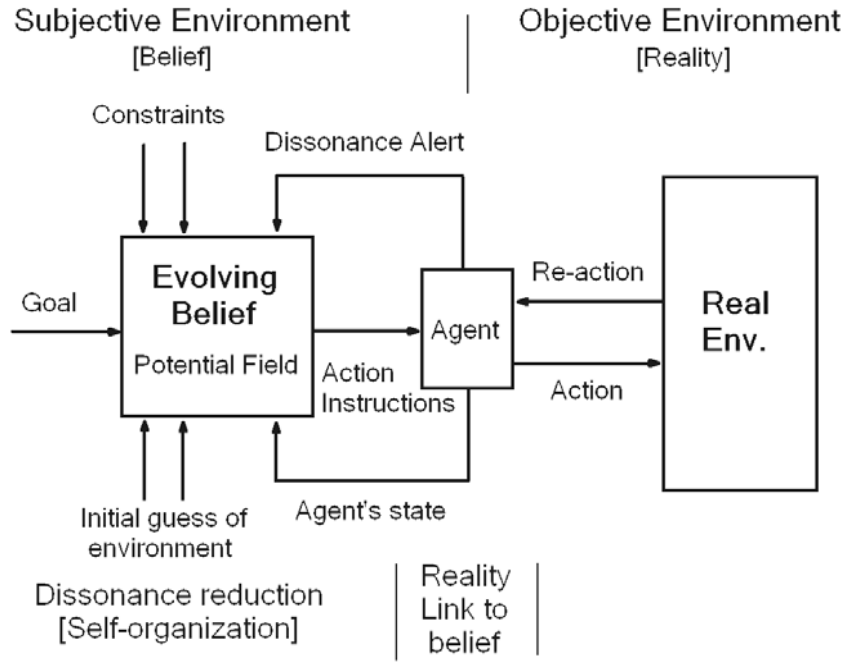

Figure-7: The structure of an EHPC

An EHPC assumes a specific form depending on the boundary value problem (BVP) used for synthesizing the potential $V_i$. The Dirichlet BVP, shown in equation 13, is used here for generating the PRF control components:

$$\nabla^2 V_i(x) \equiv 0 \qquad\qquad x \in R^N - \Gamma\grave{}_i - C_i \qquad (13)$$

subject to: $\qquad V_i = 0|_{X=C_i} \;\&\; V_i = 1|_{X \in \Gamma\grave{}_i}$ .

A sample of the behavior generated by an EHPC using such a BVP [47] is shown in figure-8. For a proof of correctness, the reader is referred to [49].

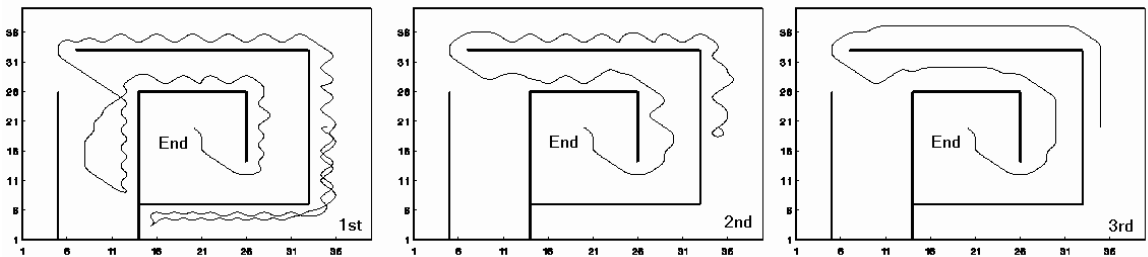

Figure-8: Three successive attempts of a point agent to navigate an unknown environment

## B. The CRF Control

There are only two ways conflict could arise in a workspace occupied by more than one purposive, mobile agent, each of which is capable of safely reaching its target in the absence of the others:

1- Two or more agents may attempt to occupy the same space at the same time.

2- Two or more agents may block each other's way preventing movement towards the targets.

A conflict resolving control (**uc**$_i$) that can prevent the above two events from happening will enable the utilizing agent to reach its target. It is obvious that an agent can prevent another from moving towards it, hence occupying the same space it is using, by exerting a force that is radial (**ucr**$_i$) to its boundary (i.e. pushing the other agent away from it, figure-9a). On the other hand, an agent can prevent others from blocking its path by exerting a force that is tangential (**uct**$_i$) to its boundary (i.e. moving out of the way, figure-9b)

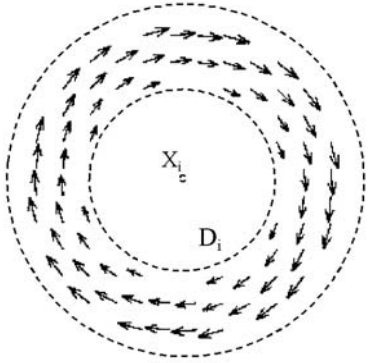 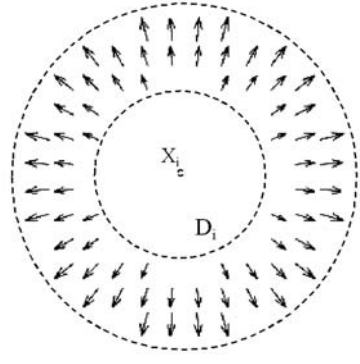

Figure-9a: Radial component of the CRF    Figure-9b: Tangential component of the CRF

The CRF component is the sum of the above two actions:

$$\mathbf{uc}_i = \mathbf{ucr}_i + \mathbf{uct}_i . \qquad (14)$$

The radial component of the control (**ucr**$_i$) may be constructed as:

$$\mathbf{ucr}_i = \sigma(|x - x_i|) \frac{\nabla Vr_i(|x - x_i|)}{|\nabla Vr_i(|x - x_i|)|} , \qquad (15)$$

where both the weighting function σ, and the scalar potentials Vr's are positive, spherically symmetric, monotonically decreasing functions whose values are zero for $|x-x_i| \geq \acute{\rho}_i$. As for **uct**$_i$, it is constructed as:

$$\mathbf{uct}_i = \sigma(|x - x_i|) \frac{\nabla \times \mathbf{A}_i(x - x_i)}{|\nabla \times \mathbf{A}_i(x - x_i)|} \qquad \nabla \cdot \mathbf{A}_i \equiv 0, \qquad (16)$$

where ∇· is the divergence operator, and $\mathbf{A}_i$ is a vector potential field [50] selected so that its gauge is zero,

$$\nabla Vr_i(|x - x_i|)^T \nabla \times \mathbf{A}_i(x - x_i) \equiv 0 \cdot \qquad (17)$$

This means that a vector potential field ($\mathbf{A}_i$) can only generate a tangent circulating field.

For the local tangent fields to form a continuous, global tangential action that has the potential to push the interacting agents out of each other's way and prevent deadlock, all the individual tangent fields must circulate along the same direction (figure-10).

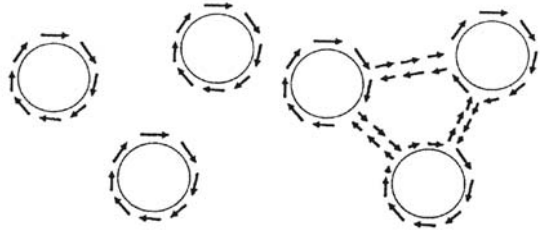

Figure-10: Same local circulations guarantee same global circulation

The overall controller governing the i'th agent is described by the dynamical system:

$$\dot{x}_i = \mathbf{ug}_i + [\mathbf{ucr}_i + \mathbf{uct}_i] + \mathbf{uo}_i \qquad (18)$$

$$= -\nabla V_i(x_i, C_i, Q_i(t_n), \hat{\Gamma}_i(t_n)) +$$

$$\sum_{\substack{j=1 \\ j \neq i}}^{K_i(t)} \sigma(|x_i - x_j|) \left[ \frac{\nabla V r_i(|x_i - x_j|)}{|\nabla V r_i(|x_i - x_j|)|} + \frac{\nabla \times \mathbf{A}_i(x_i - x_j)}{|\nabla \times \mathbf{A}_i(x_i - x_j)|} \right] +$$

$$\nabla V o_i(x_i, \hat{\Gamma}_i),$$

where $\mathbf{uo}_i = \nabla V o_i(x_i, \hat{\Gamma}_i)$, and $V o_i$ is a scalar, repelling potential field that is strictly localized to the vicinity of the obstacles. The dynamical equation governing the behavior of the collective is:

$$\begin{bmatrix} \dot{x}_1 \\ \dot{x}_2 \\ \vdots \\ \dot{x}_L \end{bmatrix} = \begin{bmatrix} \mathbf{ug}_1(x_1, C_1, Q_1(t_n), \hat{\Gamma}_1(t_n)) \\ \mathbf{ug}_2(x_2, C_2, Q_2(t_n), \hat{\Gamma}_2(t_n)) \\ \vdots \\ \mathbf{ug}_L(x_L, C_L, Q_L(t_n), \hat{\Gamma}_L(t_n)) \end{bmatrix} + \begin{bmatrix} \mathbf{uo}_1(x_1, \hat{\Gamma}_1(t_n)) \\ \mathbf{uo}_2(x_2, \hat{\Gamma}_2(t_n)) \\ \vdots \\ \mathbf{uo}_L(x_L, \hat{\Gamma}_L(t_n)) \end{bmatrix} + \begin{bmatrix} \sum_{\substack{j=2, \\ j \neq 1}}^{K_1(t)} \mathbf{uc}_1(x_i - x_j) \\ \sum_{\substack{j=1, \\ j \neq 2}}^{K_2(t)} \mathbf{uc}_2(x_i - x_j) \\ \vdots \\ \sum_{\substack{j=1, \\ j \neq L}}^{K_L(t)} \mathbf{uc}_L(x_i - x_j) \end{bmatrix} \qquad (19)$$

## V. Motion Analysis

A detailed proof of the ability of the agents, individually, to reach their respective destinations in an unknown cluttered environment may be found in [49]. While it is not hard to guarantee that the robots avoid collision with each other and with the obstacles by making the barrier controls ($\mathbf{uo}_i$, $\mathbf{ucr}_i$) excessively strong (some techniques set the strength of the control to infinity at the inner boundary of the robots [57]), their ability to converge to their respective destinations, as a group, needs careful examination. In the following it is shown that the first order dynamical systems in (18) are capable of driving the robots from anywhere in the workspace to their respective destinations provided that the narrowest passage in the workspace is wide enough to allow the largest two robots to pass at all times (i.e. no tight passages are allowed).

A. Proof of Convergence:

Here, it is shown that under certain conditions the solution of the system in (19) is globally, asymptotically stable. The proof is dependent on a theorem by LaSalle (Theorem-3, [58], pp. 524]. The theorem is restated below with minor changes to the notations.

**Theorem:** Let $\Xi(X)$ be a scalar function with continuous first partials with respect to X. Assume that :

$$1\text{-}\ \Xi(X) > 0 \qquad \forall\, X \neq C, \qquad (20)$$

$$2\text{-}\ \dot\Xi(X) \leq 0 \qquad \forall\, X.$$

Let E be the set of all points where $\dot\Xi = 0$, and M be the largest invariant set in E. Then every solution of the system:

$$\dot X = \mathbf{H}(X, C, Q, \Psi, \Gamma\grave{\ }) \qquad (21)$$

bounded for $t \geq 0$ approaches M as $t \to \infty$.

**Proposition-1:** For the system in (19), there exists a set of **uct**'s that can guarantee

$$\lim_{t \to \infty} X(t) \to C, \qquad (22)$$

provided that:   1- for the gradient dynamical systems:

$$\dot x_i = -\nabla V_i(x_i, C_i, Q_i(t_n), \Gamma\grave{\ }_i(t_n)), \qquad (23)$$

$$\lim_{t \to \infty} x_i(t) \to C_i \qquad\qquad i=1,\ldots,L$$

2-   $D_i \cap D_j \equiv \phi, \qquad i \neq j$
$D_i \cap O \equiv \phi,$

3-   $\forall\, x\grave{\ } \in \Omega$   there-exists xc, such than
$x\grave{\ } \in \{\, x : |x - xc| \leq \xi \,\} \subset \Omega$

where $\xi = \rho\grave{\ }_1 + \rho\grave{\ }_2$, where $\rho\grave{\ }_1$ and $\rho\grave{\ }_2$ are the expanded radii of the two largest robots in the group.

The third condition guarantees that nowhere in $\Omega$ will the geometry of the environment prevent the agents from resolving a conflict. The inability to resolve a conflict is the result of an agent being forced to project motion along environmentally-determined degrees of freedom (Figure-11). The forced pattern of motion may not lend itself to the resolution of the conflict.

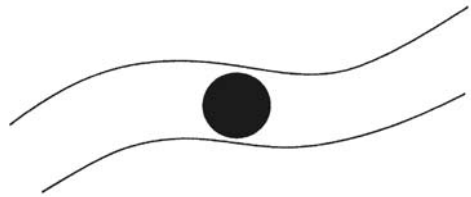

Figure-11: Restrictive environments force *a priori* determined spatial patterns on movement

Its is important to guarantee that there always exists a local, simply-connected region that is large enough to enable any two robots to interact. This ensures the realization of conflict resolution no mater what pattern of motion the agents arrive at.

**Proof:** consider the following Lyapunov function candidate:

$$\Xi(X) = \sum_{i=1}^{L} V_i(x_i), \qquad (24)$$

where $V_i(x_i)$ is used to refer to $V_i(x_i, C_i, Q_i(t_n), \Gamma\grave{\ }_i(t_n))$, and $Vo_i(x_i)$ refers to $Vo_i(x_i, \Gamma\grave{\ }_i)$. It was shown in [49] that harmonic potential fields are Lyapunov function candidates, i.e. $V_i(x_i) = 0$ for $x_i = C_i$, and $V_i(x_i) > 0$ for $x_i \neq C_i$. Therefore the above sum is a valid Lyapunov function candidate, i.e. $\Xi(X) = 0$ for $X = C$, and $\Xi(X) > 0$ for $X \neq C$. The time derivative of $\Xi$ may be computed as:

$$\frac{d}{dt}\Xi = \sum_{i=1}^{L} \nabla V_i(x_i)^t \frac{d}{dt} x_i \qquad (25)$$

$$= \sum_{i=1}^{L} \nabla V_i(x_i)^t [-\nabla V_i(x_i) +$$

$$\sum_{\substack{j=1 \\ i \neq j}}^{K_i(t)} \sigma(|x_i - x_j|) \left[ \frac{\nabla Vr_i(|x_i - x_j|)}{|\nabla Vr_i(|x_i - x_j|)|} + \frac{\nabla \times \mathbf{A}_i(x_i - x_j)}{|\nabla \times \mathbf{A}_i(x_i - x_j)|} \right] + \nabla Vo_i(x_i)].$$

The above expression is examined term by term to determine the nature of the time derivative of $\Xi$. It is obvious that the term:

$$\sum_{i=1}^{L} -\nabla V_i(x_i)^t \nabla V_i(x_i) \qquad (26)$$

is negative definite with a zero value (stable global equilibrium) at and only at $x_i = C_i$, $i=1,..,L$, ($X=C$). As for the term:

$$\sum_{i=1}^{L} \nabla V_i(x_i)^t \nabla Vo_i(x_i), \qquad (27)$$

One must first notice that $\nabla Vo$ is a local field that is strictly limited to a thin narrow region surrounding $\Gamma$. Its value is zero everywhere else in $\Omega$. By construction, the field lines of $\nabla Vo_i$ emanate normal to $\Gamma$ (in order to drive the robot away from the obstacles):

$$\nabla Vo(x_i) = \begin{bmatrix} \alpha(x_i)\mathbf{n} & x_i \in \Gamma_i^` \\ 0 & \text{elsewhere} \end{bmatrix} \qquad (28)$$

where $\mathbf{n}$ is a unit vector that is normal to $\Gamma_i^`$, and $\alpha$ is a smooth, positive, monotonically decreasing scalar function with a value set to zero a small distance ($\epsilon$) away from the boundary of the obstacles ($x_i^t\mathbf{n}$), i.e. $\alpha(x_i)=0$ for $x_i^t\mathbf{n} > \epsilon$. The BVPs used for constructing the potential field associated with the PRF control ($V_i$) admits only two types of basic boundary conditions (BCs):

1- homogeneous Neumann BCs:

$$\frac{\partial}{\partial \mathbf{n}} V_i(x_i) = \nabla V_i(x_i)^t \mathbf{n} \equiv 0, \qquad x_i = \Gamma_i^` \qquad (29)$$

2-homogeneous Dirichlet BCs: $\quad V_i(x_i) = 1,$

which in turn makes: $\qquad \dfrac{\partial}{\partial \mathbf{n}} V_i(x_i) = \nabla V_i(x_i)^t \mathbf{n} < 0, \qquad (30)$

(i.e. the maximum of $V_i$ is achieved at $x_i = \Gamma_i^`$ and its value decreases with motion away from $x_i = \Gamma_i^`$). As a result the above term is either:

$$\sum_{i=1}^{L} \nabla V_i(x_i)^t \nabla Vo_i(x_i) \equiv 0, \qquad (31)$$

or:
$$\sum_{i=1}^{L} \nabla V_i(x_i)^t \nabla Vo_i(x_i) < 0, \qquad x_i = \Gamma`_i.$$

As for the second term of (25), it ought to be mentioned that forces surrounding the mobile agents (CRFs) have a local, reactive, passive nature. In view of the above, this guarantees that no unbounded growth in the magnitude of the $x_i$'s can occur. The worst case is for those forces to cause a deadlock in motion (i.e, X - C = constant, t→∞). Since in the worst case scenario, motion will be brought to a halt (i.e, $\dot{\Xi} = 0$), also taking into consideration the negative definiteness of the other terms, the time derivative of $\Xi$ is always less than or equal to zero:
$$\dot{\Xi} \leq 0. \qquad (32)$$

If the i'th robot enters a static equilibrium before the target is reached, the following identity must hold:
$$\sum_{j=1}^{K_i(t)} \sigma(|x_i - x_j|)\left[\frac{\nabla Vr_i(|x_i - x_j|)}{|\nabla Vr_i(|x_i - x_j|)|} + \frac{\nabla \times \mathbf{A_i}(x_i - x_j)}{|\nabla \times \mathbf{A_i}(x_i - x_j)|}\right] = \nabla Vo_i(x_i) - \nabla V_i(x_i). \qquad (33)$$

Therefore, the set E is equal to: $\qquad E = E1 \cup E2 = \{x_i : \frac{d}{dt}\Xi = 0\}, \qquad (34)$

where: $\qquad E1 = \bigcup_i E1_i, \quad E1_i = \{x_i : x_i = C_i\} \qquad i=1,...,L$

and $E2 = \bigcup_i E2_i$, where

$$E2_i = \{x_i : \nabla Vo_i(x_i) - \nabla V_i(x_i) + \sum_{\substack{j=1 \\ i \neq j}}^{K_i(t)} \sigma(|x_i - x_j|)\left[\frac{\nabla V_i(|x_i - x_j|)}{|\nabla V_i(|x_i - x_j|)|} + \frac{\nabla \times \mathbf{A_i}(x_i - x_j)}{|\nabla \times \mathbf{A_i}(x_i - x_j)|}\right] = 0, \quad x_i \neq C_i\}. \qquad (35)$$

The largest invariant set M⊂E is the subset of E that satisfies the equilibrium condition on (21). Before computing M, let us first examine if E2 is an equilibrium set of system (21). For this case the system forces may be computed using the equation:

$$h_i = \nabla Vo_i(x_i) - \nabla V_i(x_i) + \sum_{\substack{j=1 \\ i \neq j}}^{K_i(t)} \sigma(|x_i - x_j|)\left[\frac{\nabla Vr_i(|x_i - x_j|)}{|\nabla Vr_i(|x_i - x_j|)|} + \frac{\nabla \times \mathbf{A_i}(x_i - x_j)}{|\nabla \times \mathbf{A_i}(x_i - x_j)|}\right]. \quad i=1,..,L. \qquad (36)$$

It should be noticed that if the second condition of (23) holds, the magnitude of the radial reaction forces ($\nabla Vo_i$, and $\nabla Vr_i$) is determined by the self-forces ($\nabla V_i$) and the geometric configuration the robots assume during deadlock. On the other hand, the magnitude of the circulating forces ($\nabla \times \mathbf{A_i}$) is totally independent of the self-forces. Since the individual circulating forces are made to rotate in the same direction, such fields contain no singularities (Figure-12). In other words, the circulating forces never vanish, always guaranteeing that relative motion among the agents can be actuated. The strength of these fields can be independently set by the designer anywhere in the workspace. Since the goal is to eliminate E2 from M, this freedom is used to guarantee that $h_i \neq 0 \; \forall \; x_i \neq C_i$, i=1,..L. In other words, the robots will always be moving whenever they are in close proximity to each other (i.e., no deadlock).

Since the self-forces are generated from the gradient flow of a harmonic potential, their magnitude in $\Omega$ is bounded:
$$|\nabla V_i(x_i)| \leq B_i, \quad x_i \neq C_i \qquad i=1,..,L, \qquad (37)$$
where $B_i$ is a positive, and finite constant. Also notice that it is not possible for the magnitude of the passive reaction forces to exceed that of the self-forces. Therefore, a simple and conservative choice of the magnitude of the circulating field that would guarantee that E2 is not an equilibrium set of (21) is:

$$|\nabla \times \mathbf{A_i}(x_i - x_i)| \geq \sum_{i=1}^{L} B_i. \tag{38}$$

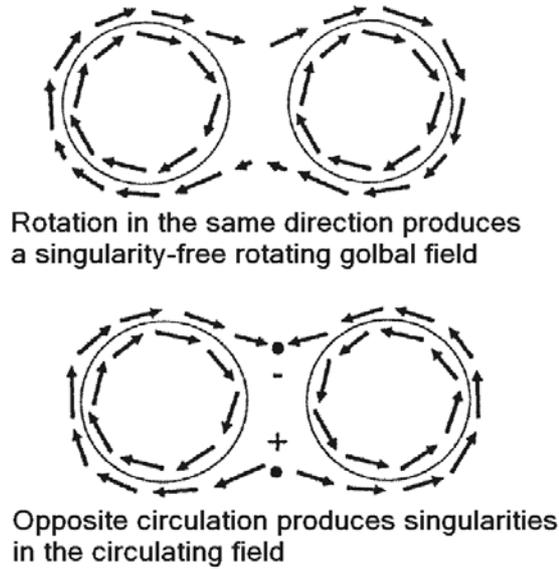

Rotation in the same direction produces
a singularity-free rotating golbal field

Opposite circulation produces singularities
in the circulating field

Figure-12: Tangent fields with same circulation are free of singularities

It should also be noticed that if the third condition of (23) is not satisfied (i.e. there is not enough free space for the largest two robots to move at all times) and the circulating fields have to push against a static obstacle (a static obstacle can exert infinite reaction force), no realizable choice of $B_i$'s would exist to satisfy condition (38). The above treatment amounts to the simple physical fact that whenever the radial reaction forces of one or more robots are in equilibrium the circulating forces intervene to pull the system out of deadlock. If the above condition is satisfied, E2 is eliminated from M. Also, since the robots have convex geometry, no equilibrium paths can form, trapping one or more robots in a limit cycle. This means that continuous motion along the tangent of a robot will eventually lead to a move away from that robot, hence resolving the conflict.

As for E1, the fact that the $T_i$'s are taken so that $D\grave{}_i \subset T_i$, guarantees that once the robots reach their respective destinations, no interactions among their fields can occur (i.e $\mathbf{uc_i} = 0$, and $\nabla Vo_i = 0$, i=1,..,L). Also since:

$$\nabla V_i(x_i) = 0, \quad x_i = C_i, \tag{39}$$

system (21) reduces to:
$$\dot{x}_i = 0, \quad x_i = C_i \tag{40}$$

making the largest invariant set equal to: $\quad M = \bigcup_i \{x_i : x_i = C_i\} \quad$ i=1,..,L. $\qquad$ (41)

Therefore, according to LaSalle's theorem, the robots will globally, asymptotically converge to their respective destinations, i.e. : $\qquad \lim_{t \to \infty} x_i \to C_i \qquad$ i=1,..,L $\qquad$ (42)

B. A Note on Completeness:

As mentioned earlier, the suggested planner is conditionally-complete provided that conditions (23) and (38)

hold. To examine why imposing the third condition of (23) is necessary for the suggested planner to guarantee completeness, note that behavior, in general, has two components: a spatial one that consists of a vector field that assigns to each point in the workspace a direction along which motion should proceed. It also has a temporal component which consists of a scalar field that assigns a speed to each point in the workspace. Therefore, completeness for a general class of workspaces implies the existence of a spatio-temporal pattern of behavior which, if executed by the agents, leads to the satisfaction of the goal. In general environments, where a solution exists provided that behavior be spatially and temporally manipulated, the environment may, at any one stage, deprive the planner of the ability to fully manipulate spatial behavior. This could happen by forcing one agent or more to follow predetermined spatial behavioral patterns that are set by the geometry of the workspace (figure-11). If such a situation occurs, the conflict can only be resolved by manipulating the temporal component of behavior (i.e speed up or slow down the movements of the agents, as well as halt motion or reverse it). Since the suggested planner is totally reliant on manipulating spatial behavior only, it may fail if it encounters situations where both spatial and temporal behavior have to be manipulated. The third condition of (23) guarantees that the environment will never be able to prevent the planner from spatially manipulated behavior in order to resolve a conflict. In a recent study by the author [49], a method for synthesizing a PRF control component that can jointly enforce regional avoidance, and directional constraints, may be used to guarantee that deadlock will not happen in environments with tight passages. Unfortunately, this approach for avoiding deadlock may reduce the set of potential solutions to the non-directionally constrained, multi-agent planning problem.

## VI. Results

Several simulation experiments were conducted to explore the behavior of the suggested method. Each case is presented as a sequence of frames with each frame depicting the state of the robots at different instants of the solution. The notation used is the same as that in the theoretical development (i.e. $D_i$ represents the i'th robot, $x_i$ its center, and $C_i$ the center of the target zone). The experiments focus on the unique capabilities of the planner, namely:

1. its ability to plan in highly congested spaces using online, sensory data only,
2. its ability to deal with unexpected events in an organizationally-closed manner,
3. its ability to tackle workspaces with dimensions higher than two as well as demonstrate the strong potential the planner has to generate dynamics-friendly trajectories.

Attempts to extend an earlier version of the work in this paper [59] were carried out in [60-63]. The primary focus of the work was on two issues: the conditioning of the differential properties of the trajectories of the agents so that they become dynamically suitable for traversal (the resulting trajectories were referred to as flyable paths), and extending the method to three-dimensional spaces. Cases 1, 2, and 3 in the following examples show that the method, in its original form, is fully capable of handling these issues.

*Case-1: A basic example:*

In figure-13 two robots sharing the same obstacle-free workspace are required to exchange positions. In doing so, each robot makes the simple, but naive, decision of moving along a straight line to the target. Despite the apparent conflict which each is heading towards, each robot proceeds with its plan as if the selected action is conflict-free. Once the conflict is in a phase that is detectable by the robot's local sensors, corrective actions are taken by each to modify their behavior in order to resolve the conflict (i.e. the CRF control component is activated). As mentioned before, the "seed" CRF activities consist of a component to prevent collision, and another to move the agents out of each other's way. Once the conflict is resolved, the behavior modification activities dissipate and guidance is fully restored to the PRFs (figure-14).

The robots are circular discs with equal radii $\rho_1 = \rho_2 = \rho = 1$. The local field region surrounding the robots is the same, $\delta_1 = \delta_2 = \delta = 1.5$. The motion of the robots is described by the motions of their centers: $x_1 = [x_1 \; y_1]^t$, and $x_2 = [x_2 \; y_2]^t$. The centers are driven by the self-controllers $\mathbf{u}_1 = [ux_1 \; uy_1]^t$, and $\mathbf{u}_2 = [ux_2 \; uy_2]^t$ respectively. The self controllers have the forms:

$$\mathbf{u}_1 = \sigma(x_1, y_1, x_2, y_2) \cdot \left[ K_r \cdot \begin{bmatrix} x_1 - x_2 \\ y_1 - y_2 \end{bmatrix} + K_t \cdot \begin{bmatrix} -(y_1 - y_2) \\ (x_1 - x_2) \end{bmatrix} \right] + K_g \cdot \begin{bmatrix} -(x_1 - xr_1) \\ -(y_1 - yr_1) \end{bmatrix}$$

$$\mathbf{u}_2 = \sigma(x_1, y_1, x_2, y_2) \cdot \left[ K_r \cdot \begin{bmatrix} x_2 - x_1 \\ y_2 - y_1 \end{bmatrix} + K_t \cdot \begin{bmatrix} -(y_2 - y_1) \\ (x_2 - x_1) \end{bmatrix} \right] + K_g \cdot \begin{bmatrix} -(x_2 - xr_2) \\ -(y_2 - yr_2) \end{bmatrix}$$

$$\sigma(x_1, y_1, x_2, y_2) = (1 + \frac{\rho_1 + \rho_2 - \sqrt{(x_1 - x_2)^2 + (y_1 - y_2)^2}}{\delta}) \cdot (\Phi(\sqrt{(x_1 - x_2)^2 + (y_1 - y_2)^2} - (\rho_1 + \rho_2)) \cdot (\Phi(\delta + (\rho_1 + \rho_2) - \sqrt{(x_1 - x_2)^2 + (y_1 - y_2)^2}) \tag{43}$$

where $K_g = 0.4$, $K_r = 2$, $K_t = 1$, $\Phi(x)$ is the unit step function, $xr_1 = [xr_1 \; yr_1]^t = [4 \; 0]^t$, and $xr_2 = [xr_2 \; yr_2]^t = [-4 \; 0]^t$, $[x_1(0) \; y_1(0)]^t = [-4 \; 0]^t$, and $[x_2(0) \; y_2(0)]^t = [4 \; 0]^t$. The overall differential system governing the behavior of the robots is:

$$\begin{bmatrix} \dot{x}_1 \\ \dot{x}_2 \end{bmatrix} = \begin{bmatrix} \mathbf{u}_1(x_1, x_2, xr_1) \\ \mathbf{u}_2(x_1, x_2, xr_2) \end{bmatrix} \tag{44}$$

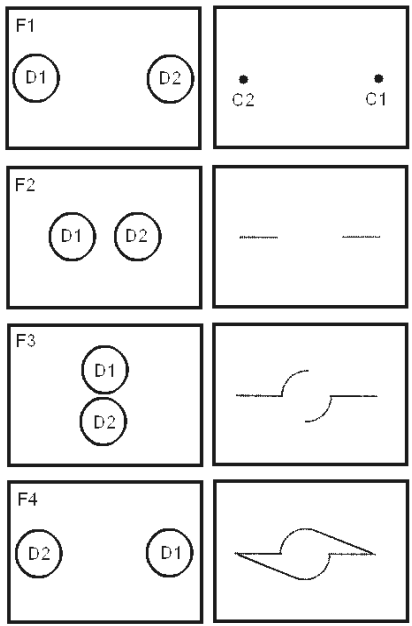

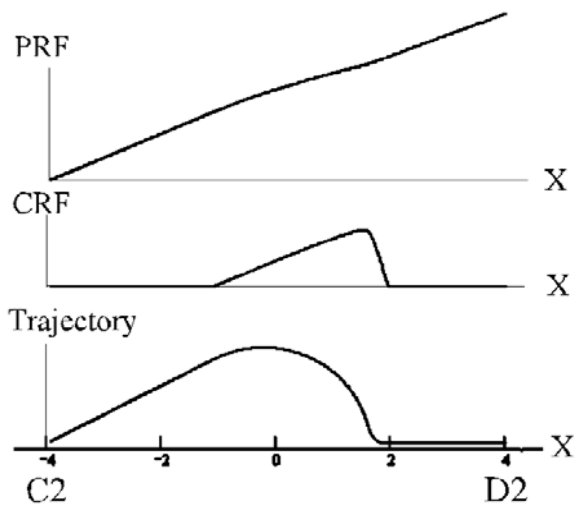

Figure-13: Two robots exchanging positions

Figure-14: CRF activities dissipate after conflict is resolved (trajectory of D2)

*Case-2: Conditioning paths' curvature:*

It is highly desirable that the generated trajectories contain as few fluctuations as possible. This is measured using the curvature $\kappa_i = d\tau_i/ds$, where $\tau_i$ is a unit vector tangent to $x_i$, and ds is an infinitesimal component of the arc. It is necessary to keep the curvature of a trajectory as small as possible if the trajectory is to be dynamically suitable for traversal. Although the focus of this paper is on generating safe trajectories for the agents to traverse to their destinations, the method has been built with dynamics in mind. There are several parameters and components of the planner that may be used to condition the differential properties of the trajectories: the weighting function σ is one of them. The following example shows that the choice of the weight profile has a pronounced effect on curvature. Three profiles are used:

linear $$\sigma(r) = (\frac{\rho - r}{\delta} + 1)(u(r - \rho) - u(r - \rho - \delta)),\qquad(45)$$

sinusoidal $$\sigma(r) = \frac{1}{2}(\cos(\frac{\pi}{\delta}(r - \rho)) + 1)(u(r - \rho) - u(r - \rho - \delta)),\text{ and}$$

exponential $$\sigma(r) = \exp(\alpha(r - \rho))(u(r - \rho)),$$

where $\alpha = \ln(\beta)/\delta$, $\beta$ is the magnitude of σ at $\rho + \delta$ ($\sigma(\rho + \delta) = \beta \ll 1$), and u is the unit step function. It ought to be mentioned that unlike the linear and sinusoidal profiles, which strictly localize that value of σ to the interval ($\rho, \rho + \delta$), the exponential profile only effectively localizes the weighting function to this period.

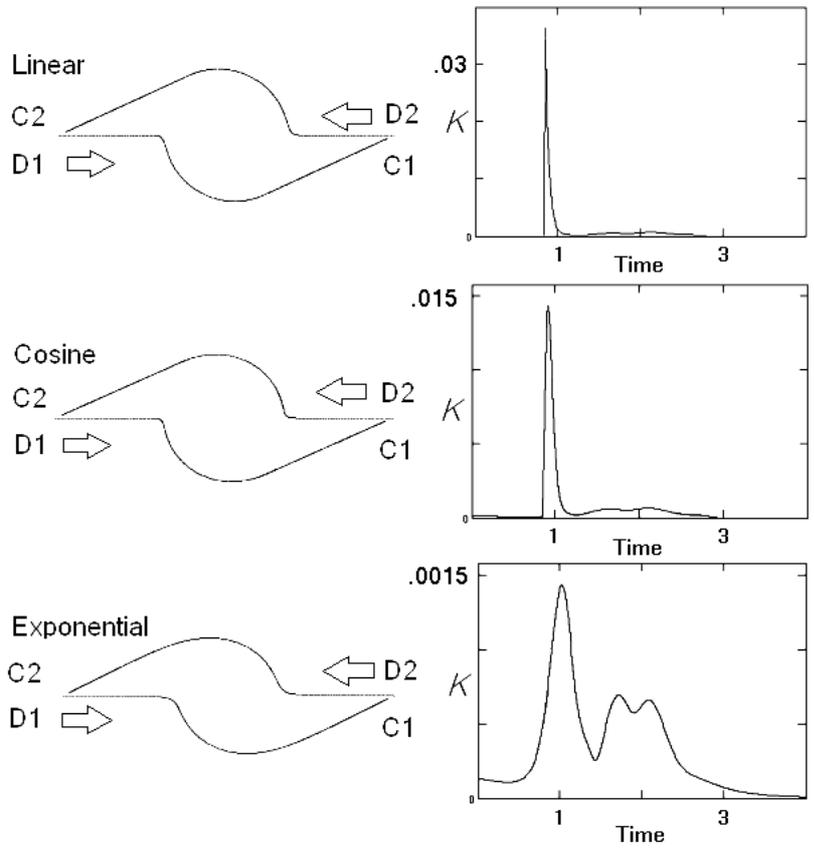

Figure-15: Effect of different CRF strength profiles on curvature

The example in case-1, which was conducted for the linear profile, is repeated for all the three profiles. The curvature of the trajectories is monitored (Figure-15). The parameters of σ are $\rho=1$, $\delta=1.5$, and $\beta=0.05$. The maximum curvature observed for the linear profile is $\kappa_{max}= 0.0363$. The maximum curvature decreased almost threefold when the smoother sinusoidal weight function was used. The maximum curvature for this case is $\kappa_{max}= 0.01434$. However, the best results were obtained for the exponential profile with a $\kappa_{max}= 0.00144$ (almost a thirty times reduction compared to the linear profile case). The above should not be considered more than a simple demonstration of the method's ability to generate dynamics-friendly trajectories. Formal investigation of this feature is left for future work.

An important parameter of the planner is the width of the action zone ($\delta$). This width must not be too small to have sharp turns. Nor should it be too large to preclude unnecessary wide deviation from the initial paths planned by the PRF fields. For all the above weight profiles, the maximum curvature is plotted as a function of $\delta$ (Figure-16). As expected, the maximum curvature is inversely proportional to $\delta$. As can be seen, a certain region for $\delta$ is reached where any increase in its value does not produce a commensurate reduction in the value of the maximum curvature. Therefore, a cutoff value for $\delta$ can be established to strike a compromise between the above two conflicting requirements.

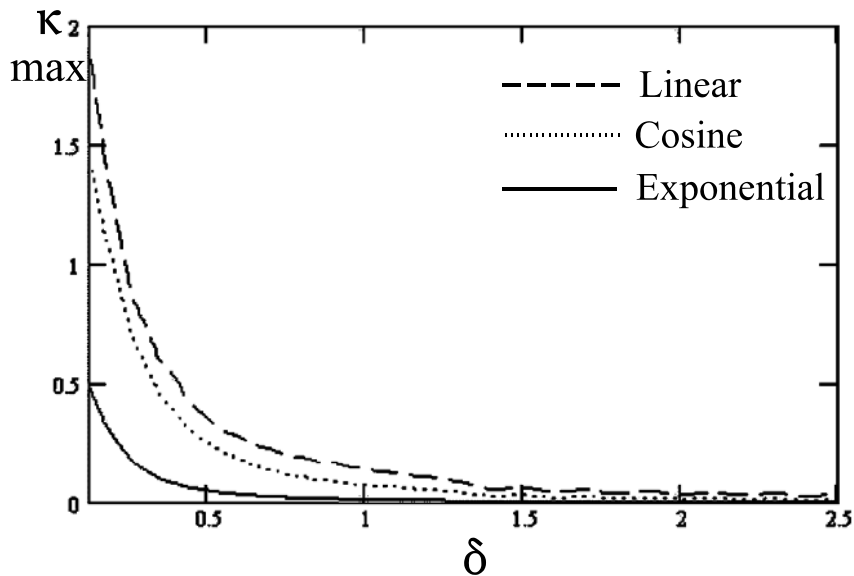

Figure-16: Maximum curvature as a function of action zone width

*Case-3: The 3-D case:*

In the sequel, all the simulation experiments are given for the 2-D case. As mentioned before, the suggested method can be applied to multi-dimensional spaces. The only reason that simulation experiments are restricted to the 2-D case has to do with the clarity of presenting the results. This is important for a qualitative understanding of the nature of the planning action the method generates. To demonstrate the applicability of the method to dimensions higher than two, the two-robot example above is repeated for the 3-D case (figure-17). Only the trajectories are plotted.

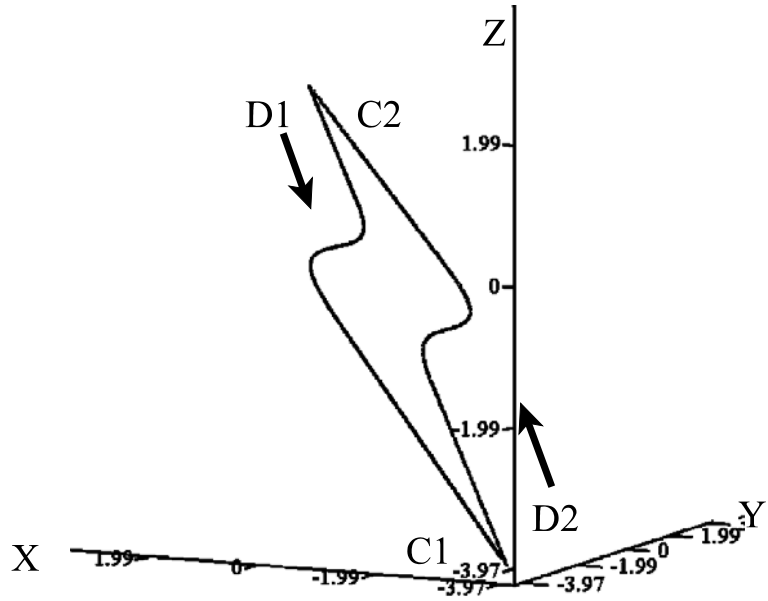

Figure-17: Two robots exchanging positions in a 3-D space

*Case-4: Fault tolerance:*

In figure-18, three robots operating in an obstacle-free space, and initially positioned on the vertices of an equilateral triangle are required to proceed towards their symmetric targets. Each robot chooses to proceed along a straight line to its target ignoring the apparent conflict to which this choice leads. For this case the response of the robots, once a conflict is detected, exhibits an interesting emergent nature. By reducing the degrees of freedom of the system from six to one, the three robots act as one rotating body to position themselves where each can proceed unimpeded towards its target. It is interesting to note that without being *a priori* programmed to do so, the robots choose to cooperate in order to resolve the conflict. This cooperation is manifested as a reduction in the degrees of freedom of the system during the period of the conflict. In a centralized system the supervisory control assigns each agent the duties it has to fulfill for the whole group to avoid conflict. If one agent fails to fulfill its obligation towards the group, the whole group may be affected. In decentralized systems, conflict evasion has a lucid nature where conflict evasion activities dynamically shift from the unable, or unwilling agents, to the remaining functional agents. Here, an agent's role keeps adapting to the situation in a manner that would, to the best of the agent's ability, enable all the agents (this includes the offending agents) to reach their targets. The following example examines this interesting property of decentralized systems. In Figure-19, a setting similar to the one in Figure-18 is used. The only difference is that D2 refuses to participate in conflict resolution and, instead, follows the plan encoded by its PRF requiring it to move along a straight line to its target. As can be seen, the remaining two agents adjust their behavior to compensate for the intransigence of D2 in such a manner that allows all the agents to reach their destinations.

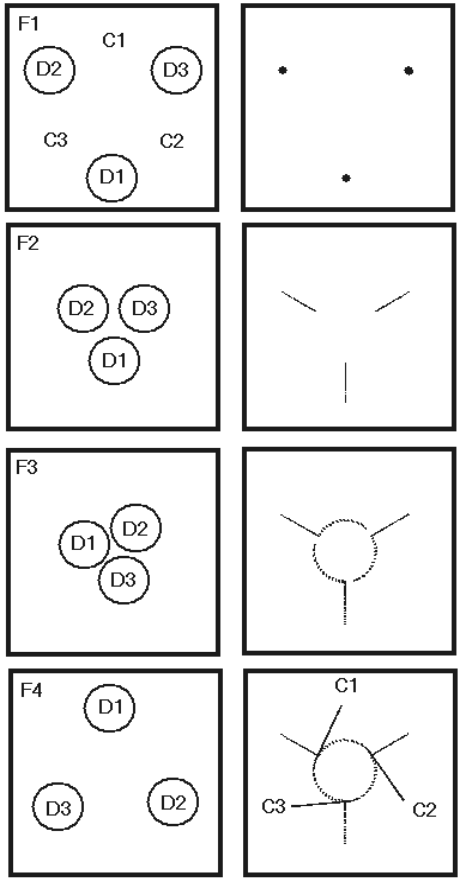
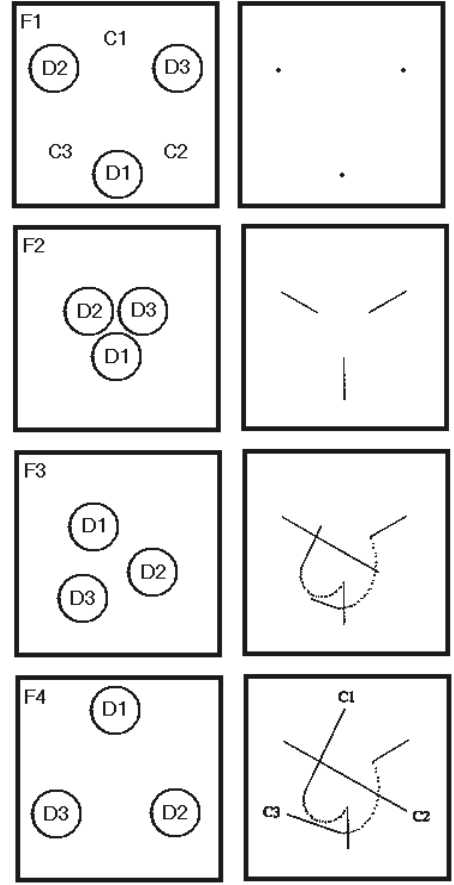

Figure-18: Three robots moving to their goals, all functioning

Figure-19: Three robots moving to their goals, D2 malfunction

*Case-5: Self-Organization:*

In the following two examples the evolutionary, cooperative, self-organizing nature of the controller is clearly demonstrated. In Figure-20 two groups of four robots each are moving in opposite directions along a road with side rails blocking each other's way. The goal is for the left group to move to the right side, and the right group to move to the left side. The groups collectively solve the problem by forming right and left lanes and confining the motion of each group to one of the lanes. It should be noted that lane formation is a high-level, holistic organizational activity that fundamentally differs from the local capabilities with which each robot is originally equipped. All eight agents are assumed to be identical with radius $\rho=1$ and local field region width $\delta=0.2$. The motion of an agent is described by the motion of its center: $x_i=[x_i \ y_i]^t$, $i=1,..,8$. The centers are directly driven by the self-controllers: $\mathbf{u}_i=[ux_i \ uy_i]^t$, $i=1,..,8$. A self-controller has the form:

$$\mathbf{u}_i = \sum_{\substack{j=1 \\ j \neq i}}^{8} \sigma(x_i, y_i, x_j, y_j) \cdot \left[ \mathbf{K_r} \cdot \begin{bmatrix} x_i - x_j \\ y_i - y_j \end{bmatrix} + \mathbf{K_t} \cdot \begin{bmatrix} -(y_i - y_j) \\ (x_i - x_j) \end{bmatrix} \right] + \mathbf{uo}(x_i, y_i) + \mathbf{ug_i}(x_i, y_i) \quad (46)$$

where σ has a form similar to the one in (43), $\mathbf{uo}(x_i,y_i)=30 \cdot [\ 0 \quad (-y_i-2)\cdot\Phi(-y_i-2)-(y_i-2)\cdot\Phi(y_i-2)]^t$, $\mathbf{ug_i}(x_i,y_i)=[\ -1 \ 0]^t$, $i=1,..,4$, $\mathbf{ug_i}(x_i,y_i)=[\ 1 \ 0]^t$, $i=5,..,8$, $K_r=20$, $K_t=10$, and $[x_1(0) \ y_1(0) \ x_2(0) \ y_2(0) \ x_3(0) \ y_3(0) \ x_4(0) \ y_4(0) \ x_5(0) \ y_5(0) \ x_6(0) \ y_6(0) \ x_7(0) \ y_7(0) \ x_8(0) \ y_8(0)]^t = [2 \ 1.3 \ 5 \ 1.3 \ 2 \ -1.3 \ 5 \ -1.3 \ -2 \ 1.3 \ -5 \ 1.3 \ -2 \ -1.3 \ -5 \ -1.3]^t$.

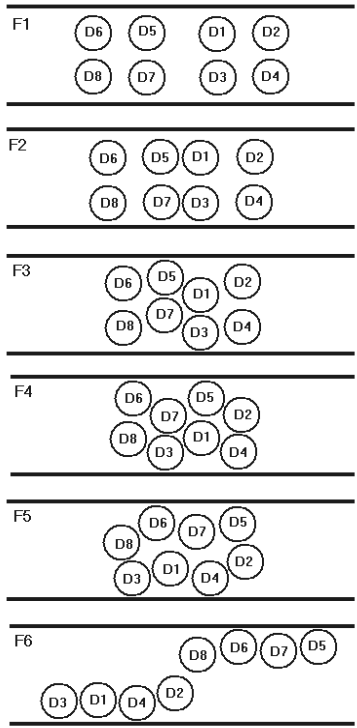
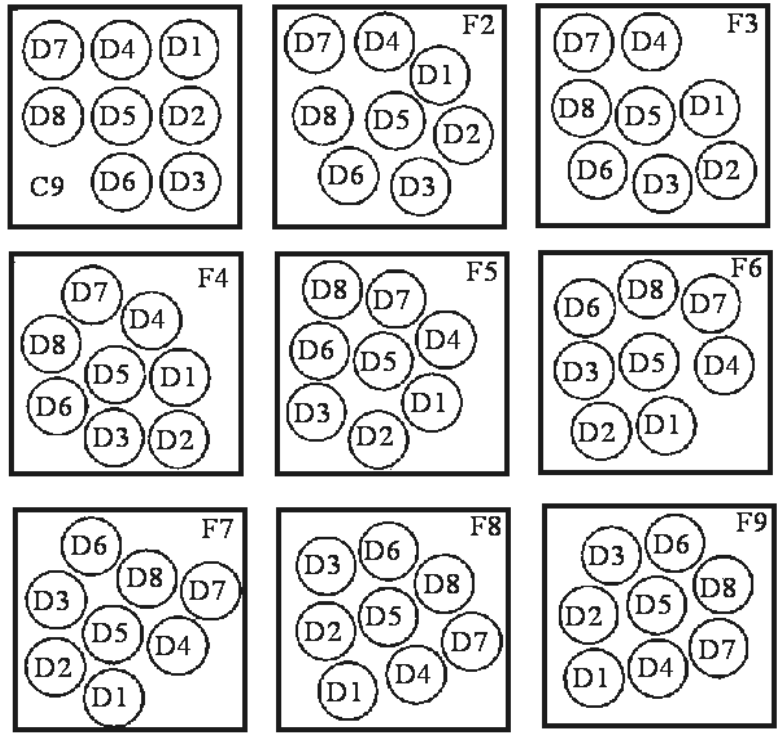

Figure-20: Two groups of robots passing each other in a confined space

Figure-21: A group of robots self-organizing to allow D1 to reach C9

In Figure-21 eight robots are confined in a box with very little room to move. The goal is for D1 to move to C9. The robots collectively reach a solution that efficiently utilizes free space. The robots solve the problem by keeping the center robot stationary, with the remaining robots rotating around it until D1 reaches its target.

*Case-6: CRF field strength and deadlock prevention:*
In the following example the importance of the circulating fields for conflict resolution is demonstrated. Here a group of eight agents is required to hold its position, except for D8 which is required to move to C8. No circulating fields are used in figure-22. As can be seen, while D8 managed to pass the first group of agents, it became trapped in a deadlock formation when it attempted to pass the second group. In figure-23 circulating fields are added. As can be seen D8 is able to reach its target, and the remaining agents maintain their original positions.

*Case-7: Planning in unknown environments:*
In Figure-24 two robots are required to exchange positions. The robots are not *a priori* aware of each other or of their surroundings. The only information they have prior to initiating action is their target locations. While the PRF fields in the previous examples are built using simple behavioral primitives, here EHPCs are used to build the PRFs. As can be seen, despite each agent's total lack of knowledge about its environment or the other member sharing the space with it, each manages to successfully reach its target from the first attempt in a conflict-free manner.

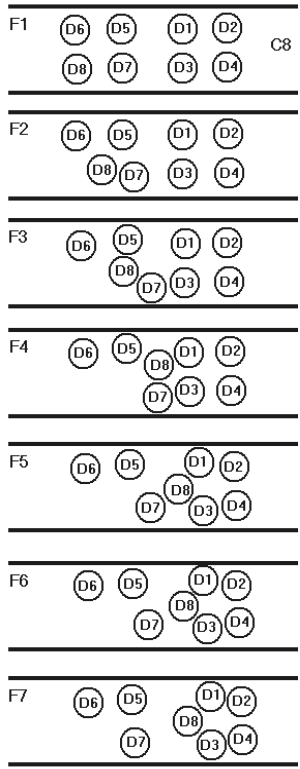
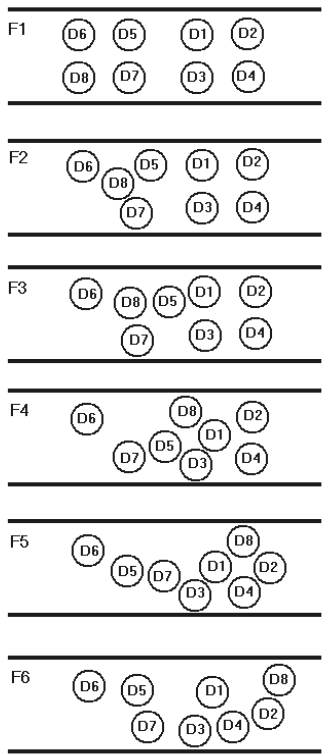
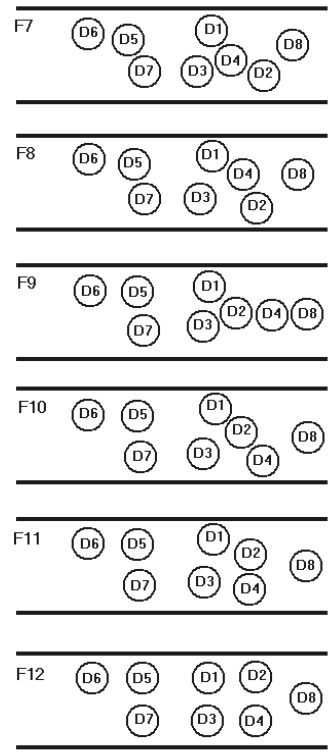

Figure-22: D1-D7 hold positions, D8 moves to C8, no Circulating fields

Figure-23: D1-D7 hold positions, D8 moves to C8, Circulating fields present.

Figure-24: Two robots exchanging positions in a cluttered, unknown environment

*Case-8: Failure with tight passages:*

While the third condition of (23) is by no means stringent, there are environments with tight passages that have only room for one robot at a time. In such a situation there are no guarantees that the multi-agent planner will function properly. Below is an example demonstrating such a situation.

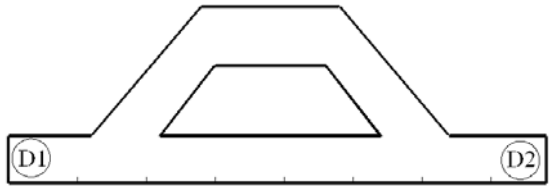

Figure-25: A workspace with tight passages

Consider the workspace in figure-25. Two robots D1 and D2 are required to exchange positions. As can be seen, the passages in Ω are not wide enough for the two robots to pass at the same time.

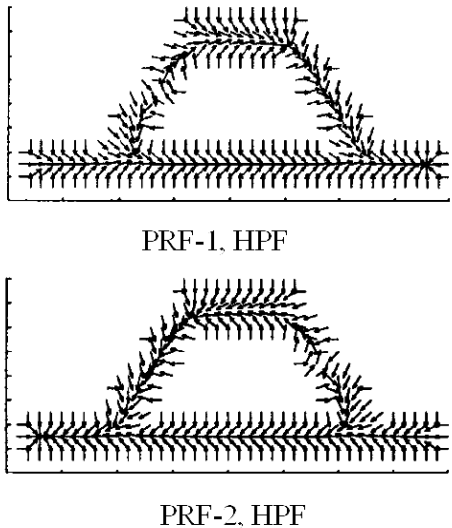

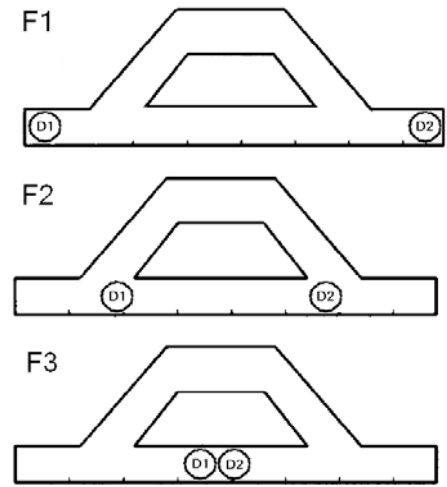

Figure-26: PRF components, HPF-based EHPC      Figure-27: Deadlock caused by a tight passage

Figure-26 shows the HPF-based PRFs for both D1 and D2. Figure-27 shows, using snapshots, the locations of the robots that are generated by the multi-agent controller at different instants of the solution. As can be seen, an unresolvable conflict arises between D1 and D2. One way to remedy this situation is to mark a tight passage as a one-way street (i.e. constrain motion in such passages to become unidirectional). This may be accomplished by using the NAHPF-based EHPC scheme. The solution, its advantages and drawbacks are discussed in [49].

**A note on complexity:**

Analysis of provably-correct, geometric, multi-agent planning methods shows that the problem has a complexity that exponentially grows in the number of agents. On the other hand, the complexity of the AL-based method suggested in this paper is linear in the number of agents enabling the method to handle planning for large groups in real-time. The reason for such a dramatic difference in complexity has to do with the action selection mechanism used by each approach. Geometric planning methods rely on search as the basis of action selection; the AL approach uses evolution instead. To construct a multi-agent AL-based controller, one has to construct a self-controller (G-type controller) for each agent individually in a manner that conforms to the AL guidelines. The multi-agent controller that is steering the group (P-type controller) evolves as a result of the interpretation of the G-type control in the context of the environment. In other words, the multi-agent controller is computed in a soft, costless manner by the process of morphogenesis. It is obvious from the above that the computational effort that has to be dispensed in constructing the multi-agent controller is equal to the sum of the effort needed for constructing the G-type controller for each agent.

## VII. Conclusions

This work has described the construction of a conditionally-complete, decentralized motion planner for agents sharing a workspace with unknown, stationary, forbidden regions. A definition for decentralization that emphasizes the autonomy of the individual agents in terms of data acquisition, information processing, and motion actuation is used as the guide for the development of the controller. The suggested multi-agent controller is found to have several attractive properties such as its ability to generate online the additional information needed to execute a successful action. It is also noted that the controller exhibits intelligent dispatching capabilities that enables it to redistribute the task of conflict evasion on the properly functioning agents. This property provides significant robustness in the case of sensor, or actuator failure. The controller employs an idea from the artificial life approach to behavior synthesis that is of central importance for the controller to achieve the above capabilities: i.e. the ability to project global useful activities through simple, local interacting activities without the agents, necessarily, being aware of the generated global behavior. The artificial life G-type and P-type control modes do support such a behavior synthesis paradigm and may be considered as the backbone for building effective decentralized controllers. The work has also presented the potential field approach as a powerful tool for generating control fields that are particularly suited to constructing intelligent, decentralized controllers.

It is important to notice that completeness of an algorithm does not exist in an absolute sense. A complete algorithm or procedure is only correct provided that certain assumptions are upheld. For example, a planner that is guaranteed to find a trajectory for a robot to a target zone may no longer be provably-correct if the implicit assumption on the path being only continuous is no longer enough (e.g., path differentiability is required). What makes a complete algorithm useful is the practicality of the conditions under which completeness is obtained. In this paper completeness is achieved provided that the linear, isotropic workspace the agents are sharing supports bidirectional movements (i.e., two-way streets). This author believes that this assumption is practical and does yield a demonstrably-useful planner.

It ought to be kept in mind that the main motive for adopting a decentralized planning strategy is to meet the stringent requirements a large-scale mobile robotics system has to satisfy in order to have a reasonable chance of success operating in a realistic environment. As the simulation results clearly reveal, a decentralized planner constructed in accordance with the AL guidelines, possesses several important properties needed, among other things, for combating the adverse effect of hardware and/or software failure likely to occur in a large scale system. They are also important in bringing the complexity of the planning task under control. In attaining these properties fundamental assumptions were made. One of these assumptions has to do with the restriction of the amount of data available for the agents to base their decision on. While this assumption is needed for operation in a decentralized mode, it may have some drawbacks. It is a well-known fact that the more information used to project an action the better is the quality of the resulting trajectory and the lower is the probability of encountering conflict situations. Moreover, in a decentralized mode, resources have to be duplicated. Each agent has to be equipped with data acquisition, data processing, decision making, and motor action modules in order to be able to operate in a decentralized mode. Compared to centralized systems where only the supervisory agent has all these faculties while the remaining agents have only relatively inexpensive motor units for executing the supervisor's commands, decentralized systems are more expensive to implement. While decentralized control is an attractive choice, in small and medium scale systems one may want to consider the centralized control option. With reliable technology, the chance of component failure is low. Moreover, with advances in computer technology, data processing and decision making algorithms, data acquisition and processing as well as planning can be done in a reliable and fast manner.

The author believes that the multi-agent controller prototype suggested in this paper will serve as a good basis for developing other multi-agent controllers. Future work will focus on conditioning the differential properties of the generated trajectories, incorporating dynamics, and generalizing the shape of the agents from that of a simple sphere to more general shapes.

**Acknowledgment:**

The author would like to thank KFUPM for its support of this work.

# Nomenclature

| | |
|---|---|
| AL | : artificial life |
| G-Type | : geno-type of behavior |
| P-Type | : pheno-type of behavior |
| EHPC | : evolutionary, hybrid, pde-ode controller |
| PRF | : purpose field component of the multi-agent controller |
| CRF | : conflict resolving field component of the multi-agent controller |
| HAS | : hyper action space |
| $V_i$ | : self-component of the potential field whose gradient is used to guide the i'th agent to its target |
| $Vr_i$ | : a potential field whose gradient forms a force field that fences the i'th agent in order to prevent collision with other agents |
| $Vo_i$ | : a potential field whose gradient field enhances the i'th agent's ability to prevent collision with stationary obstacles |
| $\mathbf{A_i}$ | : a vector potential field assigned to the i'th agent in order to generate the circulating tangent field |
| $\nabla$ | : gradient operator |
| $\nabla^2$ | : laplace operator |
| $\nabla \cdot$ | : divergence operator |
| $\nabla \times$ | : curl operator |
| $\mathbf{u_i}$ | : self-control of the i'th agent |
| $\mathbf{uo_i}$ | : stationary, obstacle avoidance component of $\mathbf{u_i}$ |
| $\mathbf{ug}_i$ | : PRF component of $\mathbf{u_i}$ |
| $\mathbf{uc}_i$ | : CRF component of $\mathbf{u_i}$ |
| $\mathbf{ucr}_i$ | : agent collision prevention radial component of $\mathbf{uc}_i$ |
| $\mathbf{uct}_i$ | : deadlock prevention tangent component of $\mathbf{uc}_i$ |
| O | : stationary obstacles |
| $\Omega$ | : workspace |
| $\Gamma$ | : boundary of the obstacles ($\Gamma = \partial O$) |
| $\Gamma_i(t)$ | : boundary of the obstacles known to the i'th agent at time t |
| $D_i$ | : the i'th agent |
| $\grave{D}_i$ | : the expanded boundary of the i'th agent |
| $S_i$ | : sensory region surrounding the i'th agent |
| $T_i$ | : parking (target) region of the i'th agent |
| $\rho_i$ | : radius of $D_i$ |
| $\acute{\rho}_i$ | : radius of $\acute{D}_i$ |
| $\delta_i$ | : width of the $S_i$ region ($\rho_i - \acute{\rho}_i$) |
| $\sigma$ | : positive, scalar, monotonically decreasing, weighting function |
| $\phi$ | : null set |
| L | : number of agents in the workspace |
| $K_i(t)$ | : number for agents in the vicinity of the i'th agent at time t |
| $\chi_i(t)$ | : set of agents in the vicinity of the i'th agent at time t |
| $Q_i$ | : a binary variable indicating the presence of a previously unknown part of the stationary environment in the vicinity of the i'th agent |
| $t_n$ | : the time instant with discrete time index n at which the value of $Q_i$ changes from 0 to 1 |
| $x_i$ | : center point of the i'th agent at time t |
| $C_i$ | : center point of the target zone $T_i$ of the i'th agent |
| $\Xi$ | : a scalar, lyapunov function candidate |
| $\dot{\Xi}$ | : the time derivative of $\Xi$ |
| E | : the set of all points where $\dot{\Xi} = 0$ |
| M | : the largest invariant set in E |
| **n** | : a unit vector normal to $\Gamma$ |
| $\kappa$ | : curvature ($\kappa = d\tau/ds$) |
| $\tau$ | : a unit vector tangent to a trajectory |